\documentclass{article} 
\usepackage{iclr2027_conference,times}

\usepackage{amsmath,amsfonts,bm}

\def\eqref#1{equation~\ref{#1}}

\def\1{\bm{1}}

\DeclareMathAlphabet{\mathsfit}{\encodingdefault}{\sfdefault}{m}{sl}
\SetMathAlphabet{\mathsfit}{bold}{\encodingdefault}{\sfdefault}{bx}{n}

\usepackage{hyperref}
\usepackage{url}
\usepackage[utf8]{inputenc} 
\usepackage[T1]{fontenc}    
\usepackage{booktabs}       
\usepackage{amsfonts}       
\usepackage{nicefrac}       
\usepackage{microtype}      
\usepackage{xcolor}         
\usepackage{graphicx}
\usepackage{caption}
\usepackage{amssymb}
\usepackage{amsmath}
\usepackage{bm}
\usepackage{tabularx}
\usepackage{wrapfig}
\usepackage{enumitem}

\usepackage{pifont}
\newcommand{\cmark}{\ding{51}}
\newcommand{\xmark}{\ding{55}}
\newcommand{\zsl}[1]{\textcolor{black}{#1}}
\newcommand{\lijie}[1]{\textcolor{black}{#1}}

\definecolor{myblue}{RGB}{10,114,188}
\definecolor{myred}{RGB}{210,32,8}
\definecolor{redbox}{RGB}{255,0,0}

\newsavebox{\hbafbox}
\sbox{\hbafbox}
{\includegraphics[width=0.49\textwidth]{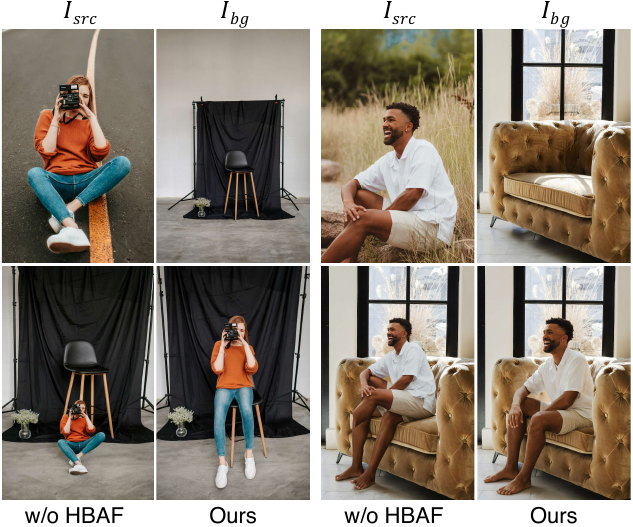}}

\usepackage{listings}
\title{InsHuman: Towards Natural and Identity-Preserving Human Insertion}

\author{%
  Jie Li$^{1,*}$, 
  Shulian Zhang$^{1,*}$, 
  Wenbo Li$^{2}$, 
  \textbf{Jian Chen}$^{1,\dagger}$, 
  \textbf{Yong Guo}$^{1,\dagger}$ 
  \thanks{Equal contribution, $^\dagger$Corresponding authors (\texttt{ellachen@scut.edu.cn}, \texttt{guoyongcs@gmail.com})} \\
  $^{1}$South China University of Technology \
  $^{2}$Chinese University of Hong Kong \
}

\iclrfinalcopy 
\begin{document}

\maketitle

\begin{abstract}
Human insertion aims to naturally place specific individuals into a target background. 
Although existing image editing models may have such ability, they often produce failure cases, including inappropriate human pose in new background, inconsistent number of people, and modified facial identity. Moreover, publicly available human datasets often lack full-body portraits and realistic physical interaction between humans and their background. 
To address these challenges, we propose \textbf{InsHuman} for natural and identity-preserving human insertion. Specifically, we propose \textbf{Human-Background Adaptive Fusion (HBAF)}, which detects foreground humans to obtain a binary mask and applies region-aware weighting to align the human regions between predicted and ground-truth latents, ensuring the person's pose, count, and overall appearance are coherently adapted to the target background.
We further propose \textbf{Face-to-Face ID-Preserving (FFIP)}, which detects and matches faces between the generated image and the source image in terms of face recognition features to enforce identity consistency for each face.
In addition, we propose \textbf{Bidirectional Data Pairing (BDP)} strategy to construct \emph{BDP-InsHuman}, a high-quality dataset with realistic human-background interactions. 
Experiments demonstrate that InsHuman achieves significant improvements in generating plausible images while keeping human identity unchanged.
\end{abstract}

\begin{figure}[!htb]
  \centering
  \includegraphics[width=\textwidth]{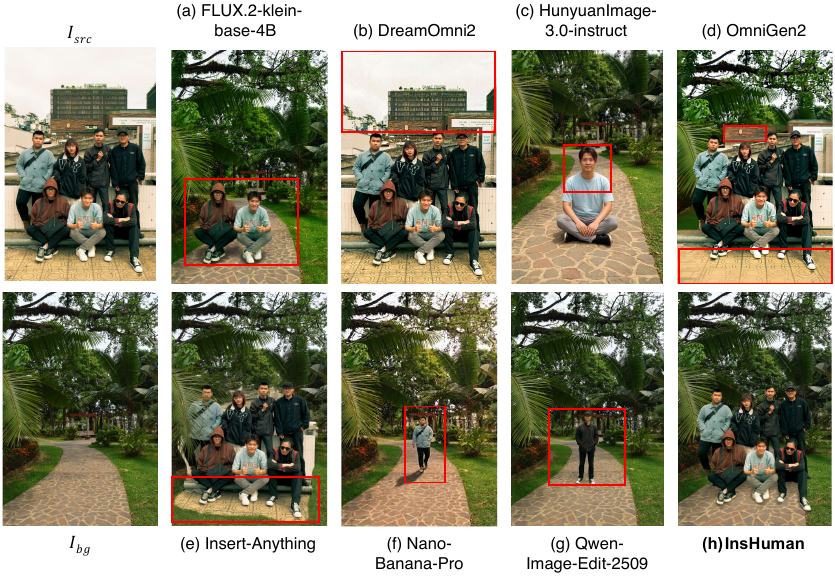}
  \caption{{\bf Qualitative comparison of different image editing models on human insertion task.} Existing image editing models often exhibit issues e.g., poses failing to adapt to the background, deviations in the number of people or appearance from the reference image, and loss of facial features. In contrast, models fine-tuned with our method (InsHuman) more stably maintain human structure and identity features, achieving more natural human insertion. \textbf{\textcolor{redbox}{Red}} boxes highlight the errors.
  }
  \label{fig:visual1}
  \vspace{-1.5em}
\end{figure}

\section{Introduction}
\label{sec:intro}
In recent years, image editing technology has made progress \citep{rombach2022high, peebles2023scalable, brooks2023instructpix2pix, hertz2022prompt}, and the image editing models \citep{flux2,xia2025dreamomni2,cao2025hunyuanimage,wu2025omnigen2,wu2025qwen}
are widely used in scenarios e.g., post-production for film and television, creative ad generation, virtual try-on \citep{choi2021viton}, social media content creation, and digital marketing. By simply providing reference images and text prompts, users can perform various complex edits e.g., background replacement \citep{yang2023paint}, style transfer \citep{tumanyan2023plug}, object rearrangement \citep{chen2024anydoor, song2023objectstitch}, and content reconstruction. Among these tasks, naturally inserting people into a target scene is becoming a practically valuable, yet technically challenging direction \citep{lu2023tf, kulal2023putting, gao2025teleportraits, zhang2026insert}. Whether it is inserting persons into a travel landscape to create immersive souvenir photos, seamlessly inserting actors into film scenes for low-cost reshoots, or placing models in different environments for advertising design, these \emph{Human Insertion} tasks require a reasonable spatial relationship between the person and the background, while preserving identity information e.g., the number of people and facial features \citep{ruiz2023dreambooth, gal2022image}. Essentially, it represents a comprehensive test of image editing models' capabilities in image understanding, spatial modeling, and identity preservation.

Directly applying existing image editing models to human insertion faces three core challenges. \textbf{First}, these models lack person-scene physical interaction priors~\citep{zhu2022celebv}, often producing poses mismatched with the background, incorrect person counts, distorted limbs, or altered clothing. 
\textbf{Second}, these models exhibit a strong "repainting" tendency, generating visually plausible but identity-inconsistent faces; while recent identity-preserving methods~\citep{ye2023ip, wang2024instantid, li2024photomaker, xiao2023fastcomposer} have made progress, balancing facial fidelity with natural scene integration remains unsolved. 
\textbf{Third}, existing portrait datasets (e.g., FFHQ~\citep{karras2019style}, VGGFace2~\citep{cao2018vggface2}, as well as generic human datasets~\citep{liu2016deepfashion, shao2018crowdhuman}) mostly consist of head or half-body shots with plain backgrounds, lacking realistic real person-scene interactions, making high-quality paired training data difficult to obtain.

Regarding these challenges, we address them in three aspects. 
\textbf{First}, to establish a reasonable interaction between the inserted human and the background, we \textbf{strengthen the model’s attention to the region of the whole human body, guiding it to prioritize learning the pose and interactive relationship with the background}. We observe that global structural information is mainly learned during the early denoising stage of diffusion models and remains relatively stable thereafter. Motivated by this, we introduce a timestep-adaptive strategy that applies stronger guidance at early stages and gradually reduces its weight as the timestep decreases, following the learning dynamics of diffusion models. 
\textbf{Second}, to preserve human identity, we introduce a facial constraint that minimizes the distance between face features extracted from the source and predicted images. While identity-preserving objectives are not new, our timestep-adaptive training \textbf{enables the model to jointly learn human pose and retain facial identity.}
\textbf{Third}, to address the lack of human insertion data, we construct a dataset containing hundreds of high-quality training pairs. Since only one of the source/background and ground-truth images can generally come from the real-world distribution while the other must be synthesized, we propose exchanging their roles, \textbf{allowing both sides to access real-world data}. This bidirectional role-exchange strategy differs from existing approaches and has been shown to be effective in practice.

In this work, we propose \textbf{InsHuman} for natural and identity-preserving human insertion.
Overall, our main \textbf{contributions} are summarized as follows:
1) We propose \emph{\textbf{Human-Background Adaptive Fusion (HBAF)}}, which adaptively adjusts the pose, position, and size of people from the source image to fit the target background while keeping the number of people unchanged. 
2) We propose \emph{\textbf{Face-to-Face ID-Preserving (FFIP)}}, a training strategy that encourages each person's face in the predicted image to maintain the same identity as in the source image. 
3) We propose \emph{\textbf{Bidirectional Data Pairing (BDP)}}, which constructs training pairs through complementary forward and reverse paths, ensuring real-world images are present on both the source and ground-truth sides of the training data.
Extensive experiments show that InsHuman significantly outperforms existing image editing models in pose adaptation, human body integrity, quantity control, and identity preservation (Figure \ref{fig:visual1}). 
Furthermore, our approach generalizes well to existing advanced image editing frameworks, reducing person count discrepancy by 57.14\% and yielding consistent improvements in facial identity consistency and overall insertion quality, as shown in Table~\ref{tab:ablation5.4}.

\section{Related Work}
\label{sec:Relate}

\textbf{Human Insertion.} 
\zsl{Existing image editing models~\citep{flux2,wu2025qwen,xia2025dreamomni2,cao2025hunyuanimage,wu2025omnigen2} achieved significant progress in general image editing and layout control \citep{li2023gligen, avrahami2022blended, couairon2022diffedit}, but often struggle with human insertion in two key aspects: pose-scene mismatch (often stemming from a lack of explicit human spatial priors ~\citep{DBLP:conf/cvpr/XuZLYLZFS24}) and loss of facial identity \citep{ruiz2023dreambooth, kumari2023multi}. Research dedicated to this task remains limited. ~\cite{kulal2023putting} learn scene affordances and person-scene interaction relationships to predict reasonable positions and poses, but lack high-fidelity facial feature control. \lijie{Text2Place~\citep{parihar2024text2place} further leverages scene priors from text-to-image models to optimize semantic masks for plausible human placement. However, its affordance prediction and human generation are handled as separate stages and require test-time optimization for each scene.} TelePortraits~\citep{gao2025teleportraits} improves identity preservation via attention manipulation, while Insert Anyone~\citep{zhang2026insert} employs a dual-branch structure to separately model identity and appearance, but depends on additional modules and shows limited robustness. In contrast, our method jointly achieves pose-scene adaptation, structural integrity, person count control, and facial identity preservation within a unified training strategy.}

\noindent\textbf{Human-Centric Datasets.} 
Full-body datasets~\citep{liu2016deepfashion,fu2022stylegan,jiang2022text2human} generally use studio-like backgrounds, while natural-scene datasets~\citep{lin2014microsoft,shao2018crowdhuman} suffer from occlusions or limited image quality.
Meanwhile, recent compositional editing benchmarks~\citep{DBLP:conf/nips/HuangSXLL23, wang2023imagen} primarily evaluate general object relations, lacking a specific focus on high-fidelity person-scene physical interactions. HumanSD~\citep{ju2023humansd} improve structural diversity but still lack paired input-output images required for scene insertion. Our \emph{BDP-InsHuman}, constructed via a bidirectional pipeline, provides realistic input-output distributions with high-fidelity person-scene interactions, offering more reliable training support.

\begin{figure}[t]
  \centering
  \includegraphics[width=\textwidth]{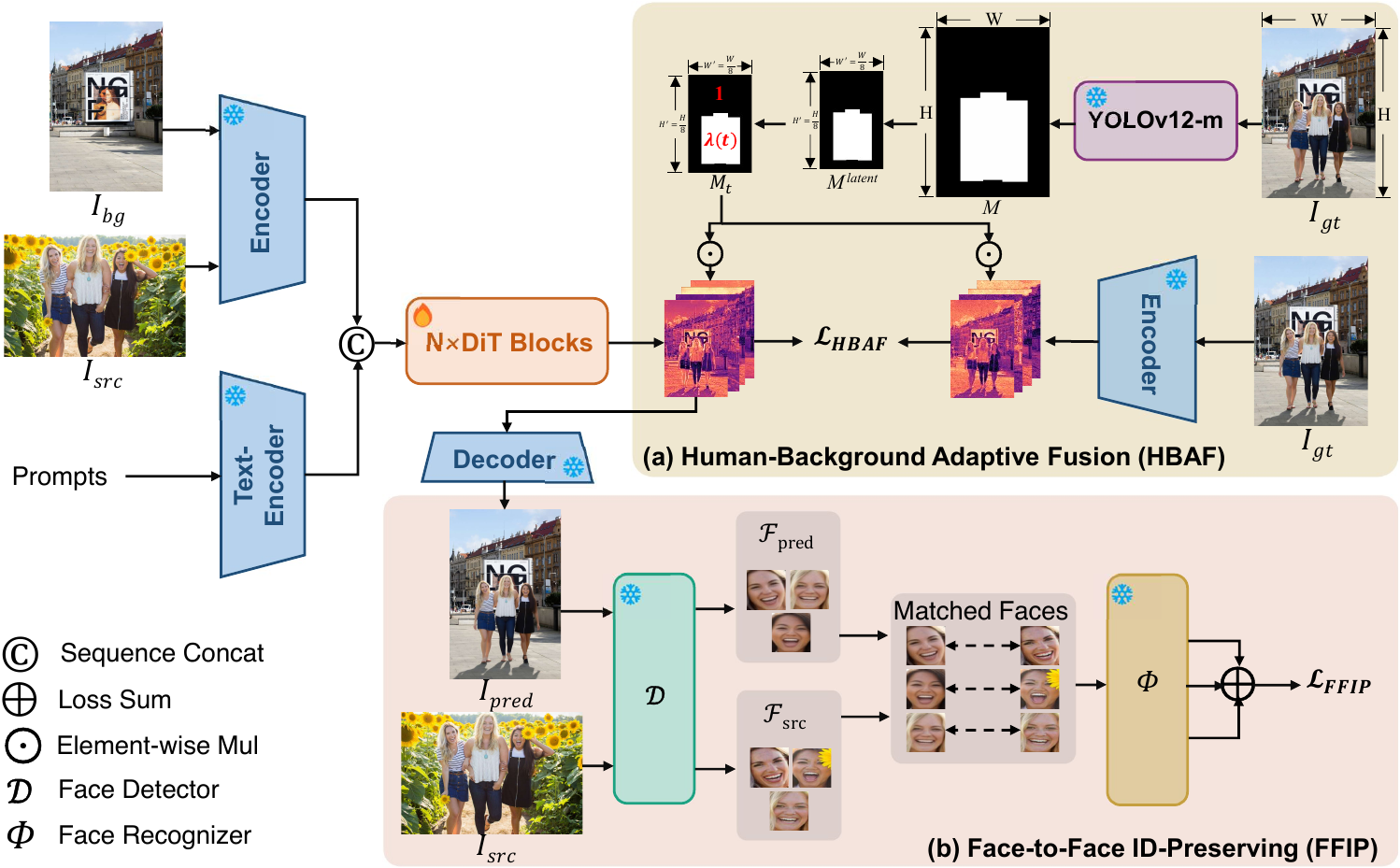}
  \caption{{\bf Overview of our proposed InsHuman.} (a) Human-Background Adaptive Fusion (HBAF): Utilizing the human region mask to dynamically adjust the weight of the human region. (b) Face-to-Face ID-Preserving (FFIP): Activated when $t \leq T_{end}$, it employs a pretrained facial feature extraction network and matching algorithm to perform facial feature alignment without interfering with the overall person structure, thereby preserving person's facial identity.}
  \label{fig:method_overview}
\end{figure}

\section{Natural and Identity-Preserving Human Insertion}
\label{sec:method}
We focus on the human insertion task, which is challenging due to three key difficulties: keeping the overall structure of the human, preserving facial identity consistency, and the lack of high-quality paired training data. To address these, we propose \textbf{HBAF} for overall human structural coherence (Section \ref{sec:HBAF}), \textbf{FFIP} for facial identity alignment (Section \ref{sec:FFIP}), and \textbf{BDP} for high-quality data construction (Section \ref{sec:BDP}). The overall framework is illustrated in Figure~\ref{fig:method_overview}



\begin{figure}[tb]
  \centering
  \includegraphics[width=\textwidth]{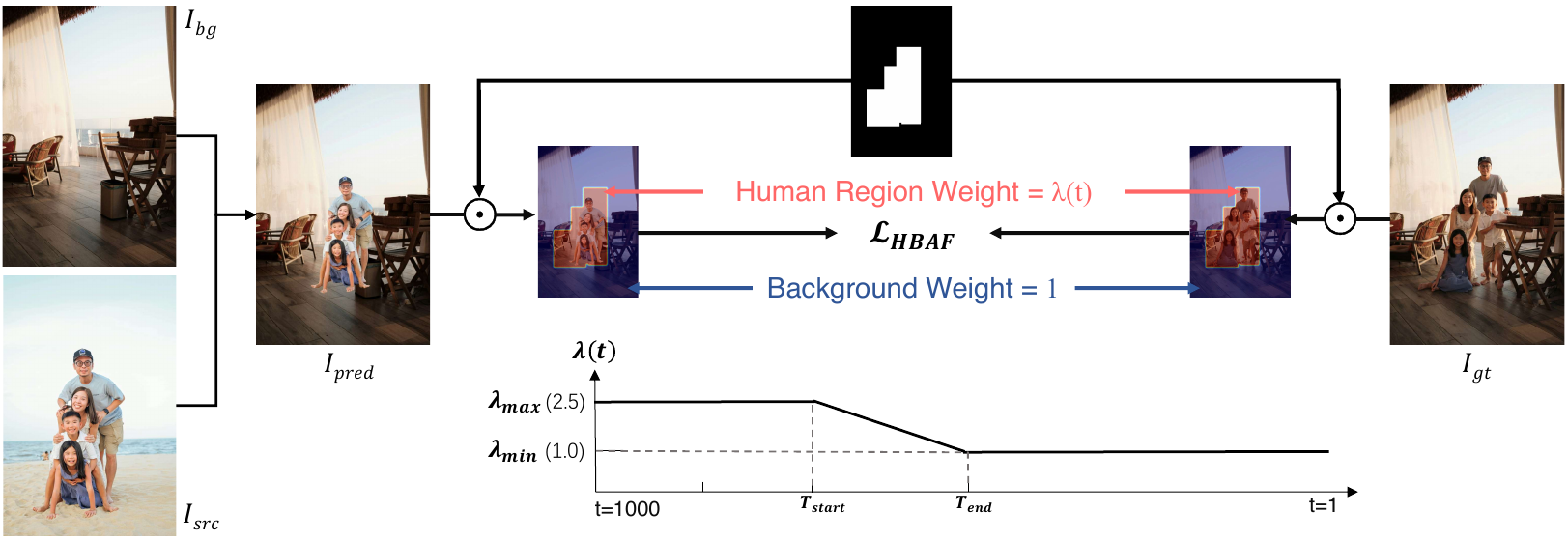}
  \caption{{\bf Weighted supervision for human region.} This mechanism uses a human region mask to guide the model in prioritizing the optimization of human structure and overall appearance.}
  \label{fig:mask-function}
  \vspace{-1em}
\end{figure}

\subsection{Human-Background Adaptive Fusion (HBAF)}
\label{sec:HBAF}
To address pose-scene mismatch, person count mismatches, and clothing inconsistencies with the reference image in human insertion, HBAF applies a region-aware weighted mask on the human area, guiding the model to prioritize structural coherence between inserted person and target background.

\begin{figure}[tb]
  \centering
  \includegraphics[width=\textwidth]{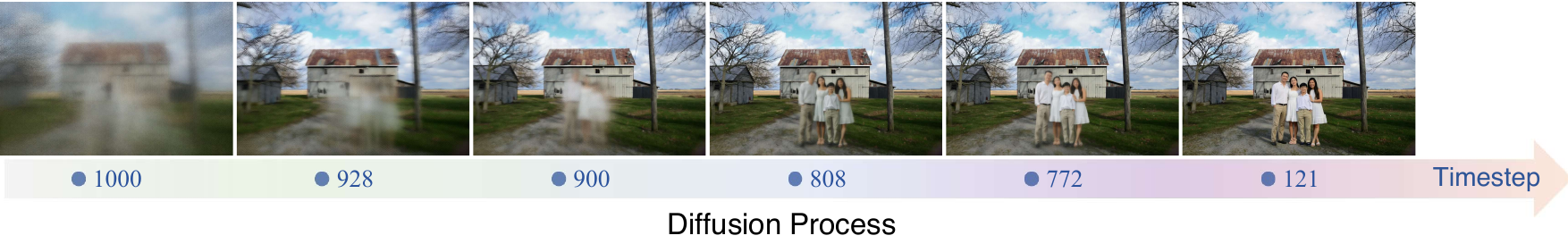}
  \caption{{\bf Variation of model-predicted images over timesteps.} In the early stages of denoising, the model primarily determines the number of people and the overall composition. In the later stages of denoising, it mainly refines details and textures.}
  \label{fig:timestep}
  \vspace{-1.5em}
\end{figure}

Given a ground-truth image $I_{gt} \in \mathbb{R}^{H \times W \times 3}$, we use YOLOv12-m~\citep{NEURIPS2025_71034442} to detect and merge all foreground persons bounding boxes, generating a binary mask $M \in \{0, 1\}^{H \times W}$, where pixels within the foreground person region are set to 1. 
As shown in Figure \ref{fig:timestep}, the model establishes overall human structure at large timesteps and refines details at small timesteps. 
We therefore design a dynamic weight function $\lambda_{t}$ that assigns a higher intensity of guidance to the human region at large timesteps and gradually relaxes it as the denoising progresses (Figure \ref{fig:mask-function}): 
\begin{equation}
  \lambda_{t} = 
  \begin{cases} 
    \lambda_{\max}, & \text{if } t > T_{start} \\[2pt]
    \lambda_{\min} + \dfrac{t - T_{end}}{T_{start} - T_{end}} 
    \cdot (\lambda_{\max} - \lambda_{\min}), & \text{if } T_{end} < t \leq T_{start} \\[6pt]
    \lambda_{\min}, & \text{if } t \leq T_{end}
  \end{cases}
  \label{eq:lambda}
\end{equation}
\lijie{$M^{latent}$ is a binary human region mask in the latent space, obtained by downsampling $M$ to the latent resolution.} We then construct a timestep-adaptive mask $M_t$ by scaling $M^{latent}$ with $\lambda_{t}$:
\begin{equation}
    M_t = \mathbf{1} + \bigl(\lambda_{t} - 1\bigr) \cdot M^{latent}
    \label{eq:adaptive_mask}
\end{equation}
where $\mathbf{1}$ is an all-one matrix. $M_t$ assigns the weight $\lambda_{t} > 1$ to the human region and the weight $1$ to the background, amplifying the supervision signal in the human area. The HBAF training loss is:
\begin{equation}
\mathcal{L}_{\mathrm{HBAF}} = \frac{1}{N} \sum_{i,j} \| M_t[i,j] \cdot (\epsilon_\theta(z_t, t, c)[:, i, j] - \epsilon[:, i, j]) \|^2
\label{eq:hbaf_loss}
\end{equation}
\lijie{$\epsilon_\theta(z_t, t, c)$ denotes the noise predicted by the model and $\epsilon$ is the ground-truth noise. $[:, i, j]$ indexes the tensor at spatial position $(i,j)$ across all channels. $ M_t[i,j]$ is the scalar weight assigned to spatial position $(i,j)$ by $M_t$. $N$ is the total number of spatial elements in the latent feature map.}
With this design, the model can better achieve adaptation between the person and the background during human insertion, while ensuring that both the number of persons and the overall appearance remain consistent with the input reference image (see Figure \ref{fig:ablation-HBAF}).

\subsection{Face-to-Face ID-Preserving (FFIP)}
\label{sec:FFIP}
Although HBAF effectively stabilizes the number of humans and the overall structure, the model still has facial distortions, resulting in inconsistencies of the facial ID between the generated image and the source image. To mitigate this, we propose Face-to-Face ID-Preserving (FFIP), which explicitly matching each face in the predicted image to its counterpart in the source image and minimizing the distance between their identity features.

We first utilize a pretrained face detection model \citep{deng2019arcface} to detect faces in both the currently predicted image $I_{pred}$ and the reference image $I_{src}$, obtaining two sets of facial regions:
$\mathcal{F}_{\text{pred}} = \{f_i^{\text{pred}}\}_{i=1}^{N_p}, \quad \mathcal{F}_{\text{src}} = \{f_j^{\text{src}}\}_{j=1}^{N_s}$,
where $N_p$ and $N_s$ denote the face counts in $I_{pred}$ and $I_{src}$, respectively. Each $f$ contains the corresponding facial bounding box information.
To establish one-to-one faces correspondences, we compute a face similarity matrix $S_{ij} = \langle \mathbf{e}_i^{\text{pred}}, \mathbf{e}_j^{\text{src}} \rangle$, where $\mathbf{e}$ is the normalized feature vector from the face recognition network and $\langle \cdot, \cdot \rangle$ denotes cosine similarity. We then apply the Hungarian algorithm to obtain the optimal face pairs $\mathcal{M}$:
\begin{equation}
    \mathcal{M} = \{(i, j) \mid j = \pi^*(i)\},  
    \pi^* = \arg \max_{\pi} \sum_{i=1}^{N_p} S_{i, \pi(i)}
\end{equation}
where $\pi(i)$ denotes the index of the face in $I_{src}$ that matches the $i$-th face in $I_{pred}$. For each matched pair, we feed the cropped facial regions into a pretrained face recognition network $\Phi(\cdot)$ with fixed parameters to extract identity features. Then, we obtain: $\mathbf{e}_i^{\text{pred}} = \Phi(f_i^{\text{pred}}), \quad \mathbf{e}_j^{\text{src}} = \Phi(f_j^{\text{src}})$. The face alignment loss is then defined as:
\begin{equation}
    \mathcal{L}_{FFIP} = \frac{1}{|\mathcal{M}|} \sum_{(i,j) \in \mathcal{M}} (1 - \langle \mathbf{e}_i^{\text{pred}}, \mathbf{e}_j^{\text{src}} \rangle)
\end{equation}
The total loss is defined as $\mathcal{L}_{total} = \mathcal{L}_{HBAF} + \lambda_{face} \mathcal{L}_{FFIP}$, where $\lambda_{face}$ is a balancing coefficient that keeps the two loss terms at a similar order of magnitude. This design guides the model to preserve facial identity consistent with the source image (Figure \ref{fig:ablation-FFIP}).

\subsection{Bidirectional Data Pairing (BDP)}
\label{sec:BDP}
Most existing human portrait datasets consist of the upper body or heads, often with blurred or even unrecognizable background information, making it difficult for the model to learn reasonable human pose and position information during training. To address this, we propose Bidirectional Data Pairing (BDP) and construct \emph{BDP-InsHuman}, a high-quality human-scene interaction dataset where each training sample consists of a source image $I_{src}$, a target background $I_{bg}$, and a ground-truth $I_{gt}$.

\begin{figure}[tb]
  \centering
  \includegraphics[width=1.0\textwidth]{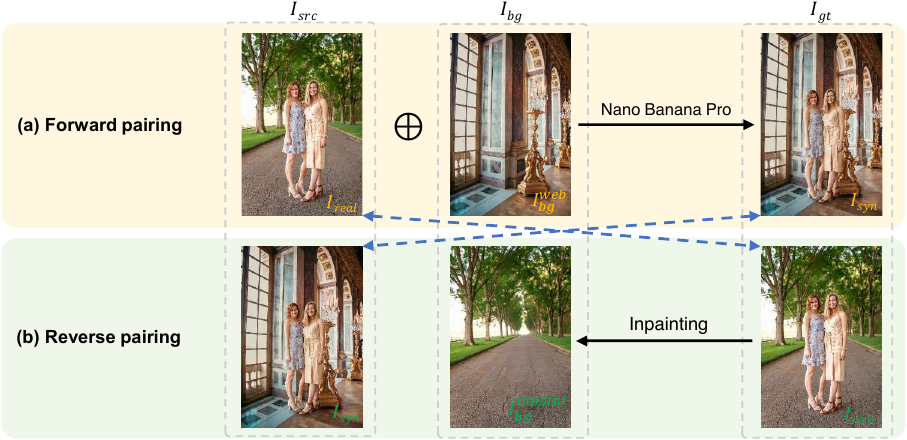}
  \captionof{figure}{{\bfseries The construction process of \emph{BDP-InsHuman}.} Through (a) forward pairing and (b) reverse pairing, real-world images can serve as both ground truth and input to the model.}
  \label{fig:BDP-InsHuman construction}
  \vspace{-1em}
\end{figure}

During training, the model takes ($I_{src}$, $I_{bg}$) as input and generates an output image under fixed text prompts (provided in Appendix~\ref{supp:prompts}). The objective is to make the generated image as close as possible to $I_{gt}$ in terms of human pose, scale, and spatial relationship with the environment.
As shown in Figure  \ref{fig:BDP-InsHuman construction} (a), in the forward pairing path, we use real-world human images $I_{real}$ \lijie{(collected from Open-source image library, discussed in supplementary)} as $I_{src}$ and web-collected scene images $I_{bg}^{web}$ as $I_{bg}$. High-quality composite results $I_{syn}$, generated by Nano Banana Pro, are selected as $I_{gt}$ under strict quality criteria covering pose plausibility, lighting consistency, and perspective correctness.

However, relying solely on generated images as ground truth risks introducing generative style biases into the model, causing it to learn synthesis artifacts rather than true physical priors. To address this, we further design a reverse pairing path, as shown in Figure \ref{fig:BDP-InsHuman construction} (b). We directly select high-quality real-world human-scene images as $I_{gt}$, and remove the person via \lijie{inpainting (Qwen-Image-Edit-2509~\citep{wu2025qwen})} to obtain $I_{bg}^{inpaint}$. The synthetic images from the forward path are then used as $I_{src}$, forming the corresponding training pair. This ensures that both $I_{src}$ and $I_{gt}$ are grounded in real-world data, enabling the model to learn more natural human insertion.


\subsection{Advantages and Differences from Existing Methods}
\label{sec:3.4difference}
InsHuman differs from existing methods in three aspects. 
\textbf{First}, for region supervision, existing methods apply fixed weights throughout training~\citep{tu2025videoanydoor}, without distinguishing structural learning from detail refinement. In contrast, HBAF uses a dynamic weight function $\lambda_{t}$, concentrating supervision at large timesteps when global human structure is learned and gradually relaxing it at smaller timesteps. 
\textbf{Second}, for identity preservation, existing methods~\citep{ye2023ip,wang2024instantid,li2024photomaker} inject reference identity features during generation, requiring additional adapters or inference-time modules. FFIP instead achieves identity consistency through training-time supervision by matching corresponding faces with the Hungarian algorithm and minimizing their identity-feature distances, without additional inference modules. 
\textbf{Third}, for data construction, existing methods mainly use generative outputs as ground truth, which may introduce synthesis biases. BDP addresses this by exchanging the roles of source and ground-truth images, allowing both to be grounded in real-world data and enabling more natural human insertion.

\section{Experiment}
\label{sec:experiment}
\subsection{Implementation details and Metrics}
\label{supp:imp_detail}
\noindent\textbf{Implementation details.} We train our model on BDP-InsHuman (Section~\ref{sec:BDP}), which consists of 529 high-quality data pairs covering diverse identities and real-world scene backgrounds, with each identity–pose combination appearing at most twice. Using the proposed \emph{HBAF} and \emph{FFIP} strategies, we perform LoRA fine-tuning on Qwen-Image-Edit-2509~\citep{wu2025qwen}. Detailed training hyperparameters and inference configurations are provided in Appendix~\ref{supp:imp_detail}.


\noindent\textbf{Evaluation and Metrics.} 
We construct a dedicated test set of 131 image pairs, each consisting of a reference image containing humans and a target background image. We adopt established evaluation metrics from prior work to assess different aspects of human insertion. 
Following MultiHuman-Testbench~\citep{borse2026multihuman}, we use \textbf{IDS to measure facial identity consistency}. Following Text2Place~\citep{parihar2024text2place}, we use \textbf{LPIPS to evaluate background preservation}. Following TISE~\citep{dinh2022tise}, we adopt \textbf{CA (Counting Alignment) to measure whether the generated image preserves the person count of the reference image}. We further use \textbf{CLIP-IQA}~\citep{wang2023exploring}, \textbf{BRISQUE}~\citep{mittal2012no}, and \textbf{UNIQUE}~\citep{DBLP:journals/spl/TemelPA16} \textbf{for perceptual quality}, and follow TelePortraits~\citep{gao2025teleportraits} to employ \textbf{VLM-T for evaluating overall human insertion quality}. Detailed metric definitions and evaluation protocols are provided in Appendix~\ref{supp:metrics}.

\subsection{\zsl{\textbf{Comparison with State-of-the-Arts}}}
\begin{figure}[t!]
  \centering
  \includegraphics[width=\textwidth]{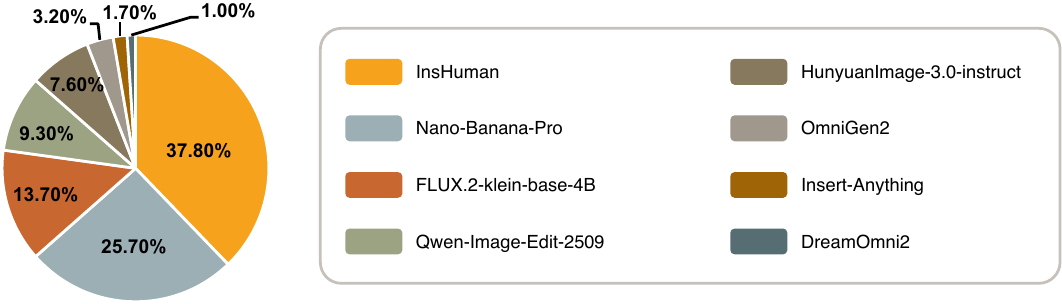}
  \caption{{\bf Human preference distribution from a blind best-pick study with 10 annotators on our test set.} InsHuman achieves the highest preference results compared to all the considered methods}.
  \label{fig:Human-preference}
  \vspace{-1.5em}
\end{figure}

\noindent\textbf{Quantitative Results.} 
InsHuman achieves the best performance across all seven metrics. Notably, our method outperforms all compared approaches on IDS (0.55), LPIPS (0.2220), CA (0.1298), three IQA metrics and VLM-T (7.8626), including both open-source models and the closed-source commercial model Nano-Banana-Pro. The human preference study (Figure~\ref{fig:Human-preference}) further validates our results, with InsHuman obtaining the highest preference rate of 37.80\%.


\begin{table}[tb]
  \caption{\textbf{Quantitative comparisons among InsHuman and other image editing models.} InsHuman achieved best performance in the following metrics.}
  \label{tab:mainTab}
  \centering
  \setlength{\tabcolsep}{1.5pt}
  \scriptsize
  \begin{tabularx}{\textwidth}{@{}l *{9}{>{\centering\arraybackslash}X}  >{\centering\arraybackslash}X@{}}
    \toprule
    \textbf{Method} & \textbf{IDS}$\uparrow$ & \textbf{LPIPS}$\downarrow$ & \textbf{CA}$\downarrow$ & {\fontsize{6pt}{7pt}\selectfont \textbf{CLIP-IQA}}$\uparrow$ & {\fontsize{6pt}{7pt}\selectfont \textbf{BRISQUE}}$\downarrow$ & {\fontsize{6pt}{7pt}\selectfont \textbf{UNIQUE}}$\uparrow$ & {\fontsize{6pt}{7pt}\selectfont \textbf{VLM-T}}$\uparrow$ \\
    \midrule
    DreamOmni2 & 0.26 & 0.5834 & 1.0229 & 0.6200 & 19.30 & 1.7205 & 4.7405 \\
    HunyuanImage-3.0-instruct & 0.21 & 0.2503 & 0.9389 & 0.6038 & 11.39 & 1.6541 & 6.7481 \\
    OmniGen2 & 0.28 & 0.4786 & 0.9542 & 0.5093 & 27.10 & 1.5245 & 6.4733 \\
    Insert-Anything & 0.28 & 0.2630 & 0.2519 & 0.4010 & 25.26 & 0.7376 & 6.7328 \\
    Nano-Banana-Pro & 0.54 & 0.2644 & 0.6336 & 0.5196 & 10.47 & 1.0436 & 7.2290 \\
    Qwen-Image-Edit-2509 & 0.50 & 0.4540 & 1.0229 & 0.6173 & 10.14 & 1.6453 & 5.9160 \\
    \midrule
    \textbf{InsHuman (Ours)} & \textbf{0.55} & \textbf{0.2220} & \textbf{0.1298} & \textbf{0.6207} & \textbf{9.50} & \textbf{1.7712} & \textbf{7.8626} \\
    \bottomrule
  \end{tabularx}
\end{table}


\noindent\textbf{Qualitative Results.} 
As shown in Figure \ref{fig:visual1}, when $I_{src}$ contains multiple people or complex poses, other models often suffer from missing persons or severe limb distortions. Comparative models also struggle to adapt the person's pose, position, and size to the target background, making the inserted person appear unnatural. 
Additional qualitative results can be found in the Appendix~\ref{supp:visual-comparison}.

\subsection{Ablation Study}
\label{sec:abla}

\begin{figure*}[t!]
  \centering
  \begin{minipage}[t]{0.5575\textwidth}
    \centering
    \includegraphics[width=\textwidth]{pic/ablation-HBAF.pdf}
    \caption{\textbf{Visual comparison without HBAF.} Removing HBAF causes the character's pose does not match the target environment and limb distortions.}
    \label{fig:ablation-HBAF}
  \end{minipage}
  \hfill
  \begin{minipage}[t]{0.4225\textwidth}
    \centering
    \includegraphics[width=\textwidth]{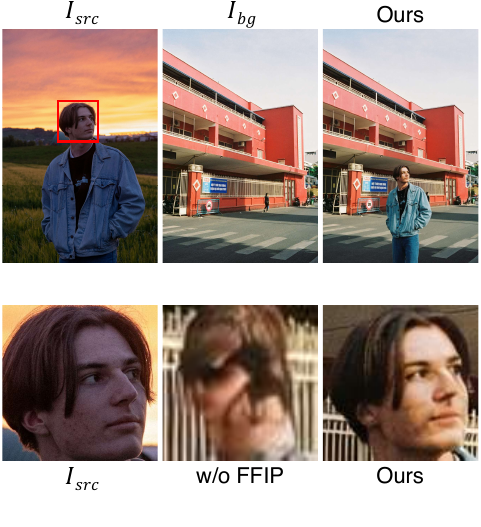}
    \caption{\textbf{Visual comparison after removing FFIP.} Rremoving FFIP, facial features differ greatly from the reference image.}
    \label{fig:ablation-FFIP}
  \end{minipage}
\end{figure*}

\noindent\textbf{Impact of HBAF and FFIP.}
As shown in Table \ref{tab:ablation5.1}, combining both HBAF and FFIP achieves the best performance. Removing HBAF leads to a significant increase in CA, accompanied by spatial mismatches and incorrect person counts (Figure \ref{fig:ablation-HBAF}). Without FFIP, facial features deviate noticeably from the reference image (Figure \ref{fig:ablation-FFIP}), confirming that both components are necessary.
\begin{table}[t]
  \centering
    \begin{minipage}[t]{0.42\textwidth}
        \centering
        \captionof{table}{\textbf{Ablation study on HBAF and FFIP.} Their combination achieves the best.}
        \label{tab:ablation5.1}
        \begin{tabularx}{\textwidth}{@{}
        >{\hsize=0.6\hsize\centering\arraybackslash}X
        >{\hsize=0.6\hsize\centering\arraybackslash}X
        >{\hsize=0.6\hsize\centering\arraybackslash}X
        >{\hsize=0.8\hsize\centering\arraybackslash}X@{}}
            \toprule
            \textbf{HBAF} & \textbf{FFIP} & \textbf{CA}$\downarrow$ & \textbf{VLM-T}$\uparrow$ \\
            \midrule
            \xmark & \cmark & 0.1985 & 7.7405 \\
            \cmark & \xmark  & 0.1445 & 7.7926 \\
            \xmark  & \xmark  & 0.2137 & 7.2115 \\
            \cmark & \cmark & \textbf{0.1298} & \textbf{7.8626} \\
            \bottomrule
        \end{tabularx}
    \end{minipage}
  \hfill
  \begin{minipage}[t]{0.54\textwidth}
    \centering
    \captionof{table}{\textbf{Ablation study on \bm{$\lambda_{\max}$} and \bm{$\lambda_{t}$} in HBAF.} Our choice of $\lambda_{\max}$ = 2.5 with dynamic $\lambda_{t}$ achieves the best performance on both CA and VLM-T.}
    \label{tab:ablation5.3}
    \begin{tabularx}{\textwidth}{@{}>{\hsize=1.7\hsize\centering\arraybackslash}X
            >{\hsize=0.6\hsize\centering\arraybackslash}X
            >{\hsize=0.7\hsize\centering\arraybackslash}X@{}}
        \toprule
        \textbf{Method} & \textbf{CA}$\downarrow$ & \textbf{VLM-T}$\uparrow$ \\
        \midrule
        $\lambda_{\max}$ = 5, w/ $\lambda_{t}$ & 0.2824 & 7.7099 \\
        Fix $\lambda$ = 2.5, w/o $\lambda_{t}$ & 0.1384 & 7.8550 \\
        $\lambda_{\max}$ = 2.5, w/ $\lambda_{t}$ (Ours) & {\textbf{0.1298}} & {\textbf{7.8626}} \\
        \bottomrule
    \end{tabularx}
  \end{minipage}
\end{table}


\begin{figure}[t!]
  \centering
  \includegraphics[width=\textwidth]{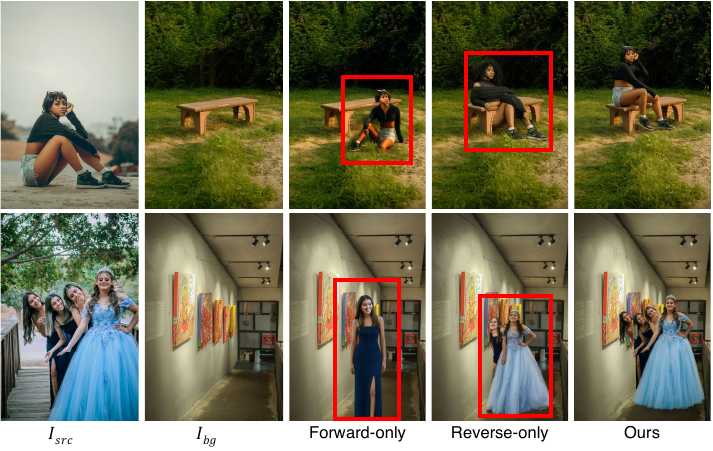}
  \caption{{\bf Ablation study results of the bidirectional construction strategy.} A comparison among forward-only construction, reverse-only construction, and bidirectional construction validates the superiority of bidirectional construction strategy.}
  \label{fig:ablation-BDP-InsHuman}
  \vspace{-1.5em}
\end{figure}

\noindent\textbf{Impact of Bidirectional Construction in BDP-InsHuman.}
As shown in Figure \ref{fig:ablation-BDP-InsHuman}, Forward-only results lack real-world physical logic between the person and background, while Reverse-only results show unnatural person size and position. In contrast, BDP achieves more natural human insertion.

\noindent\textbf{Impact of $\lambda_{\max}$ and $\lambda_{t}$ in HBAF.}
As shown in Table \ref{tab:ablation5.3}, setting $\lambda_{\max}$ at 5 causes severe structural distortions (CA=0.2824), with visual evidence provided in Appendix~\ref{supp:ablation-lambda_max} (Figure~\ref{fig:ablation-λ_max}). Fixing $\lambda$ to 2.5 without the dynamic schedule disrupts facial feature generation (as illustrated in Figure~\ref{fig:ablation-fixλ} in Appendix~\ref{supp:ablation-fix_lambda}), which confirms that both the weight magnitude and the dynamic schedule are necessary for balanced performance.

\noindent\textbf{\lijie{Impact of HBAF mask design.}}
We compare our bounding box-based mask with a finer SAM-segmented mask within the same bounding boxes. As shown in Figure~\ref{fig:supp-sam-bbox} (in Appendix~\ref{supp:ablation-SAMBbox}), the box-based mask produces more natural transitions between the inserted person and the background, since the broader supervision region encourages the model to learn perceptually important transition areas such as contact boundaries, cast shadows, and local illumination changes.

\begin{figure}[t!]
  \centering
    \includegraphics[width=\textwidth]{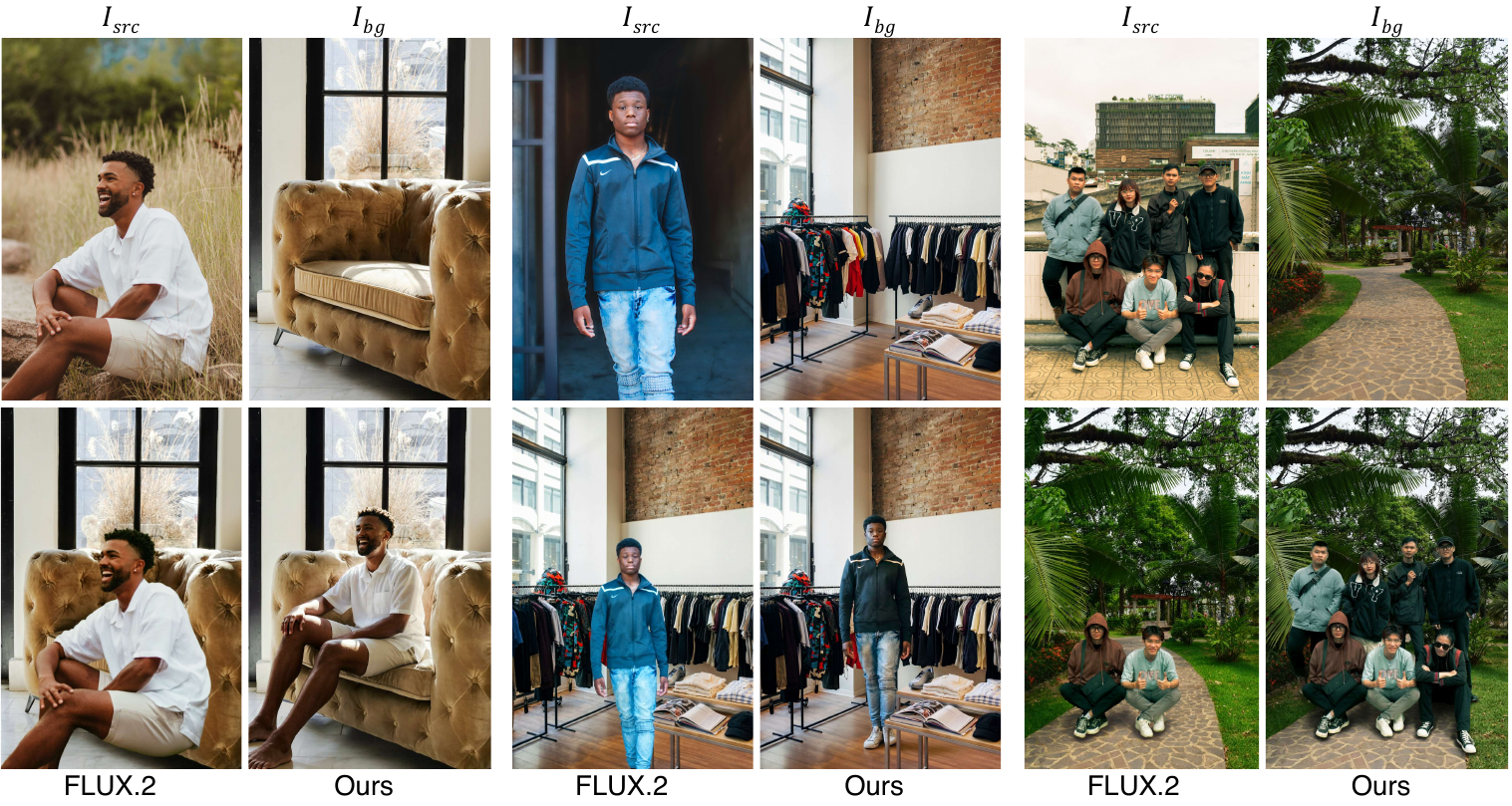}
    \captionof{figure}{\textbf{Visual comparison of FLUX.2-klein-base-4B (FLUX.2) before and after applying InsHuman.} After fine-tuning with InsHuman, the model can adaptively adjust the person's position and scale, while keeping the number of people consistent with the reference image.}
    \label{fig:ablation-flux-visual}
  \vspace{-1em}
\end{figure}

\noindent\textbf{Cross Model Generalization of InsHuman.}
To validate the generalization of InsHuman, we apply InsHuman to FLUX.2-klein-base-4B under the same settings. As shown in Table \ref{tab:ablation5.4}, consistent improvements are observed across all metrics. As illustrated in Figure~\ref{fig:ablation-flux-visual}, the fine-tuned model adapts well to diverse scenes, demonstrating that our approach is not limited to a single base model.

\begin{table}[t!]
  \captionof{table}{\textbf{Generalization of InsHuman across different image editing models.} It demonstrats that the proposed approach generalizes effectively beyond its original model and remains effective across different image editing models.}
  \label{tab:ablation5.4}
  \centering
  \begin{tabularx}{0.95\textwidth}{@{}>{\hsize=1.7\hsize\centering\arraybackslash}X
          >{\hsize=0.5\hsize\centering\arraybackslash}X
          >{\hsize=0.5\hsize\centering\arraybackslash}X
          >{\hsize=0.7\hsize\centering\arraybackslash}X
          >{\hsize=0.6\hsize\centering\arraybackslash}X@{}}
      \toprule
      \textbf{Method} & \textbf{IDS}$\uparrow$ & \textbf{CA}$\downarrow$ & \textbf{CLIP-IQA}$\uparrow$ & \textbf{VLM-T}$\uparrow$ \\
      \midrule
      Qwen-Image-Edit-2509 & 0.50 & 1.0229 & 0.6173 & 5.9160 \\
      Qwen-Image-Edit-2509 + Ours & \textbf{0.55} & \textbf{0.1298} & \textbf{0.6207} & \textbf{7.8626} \\
      \midrule
      FLUX.2-klein-base-4B & 0.61 & 0.3740 & 0.4717 & 7.3435 \\
      FLUX.2-klein-base-4B + Ours & \textbf{0.63} & \textbf{0.1603} & \textbf{0.5897} & \textbf{7.6947} \\
      \bottomrule
    \end{tabularx}
  \vspace{-1em}
\end{table}

\section{Conclusion}
\label{sec:conclusion}
We present InsHuman, a framework for natural and identity-preserving human insertion, comprising three components: HBAF, which applies a region-aware adaptive mask to ensure structural coherence between the inserted person and 
target background; FFIP, which enforces facial identity consistency by minimizing recognition-feature distances between matched face pairs; and BDP, which constructs \emph{BDP-InsHuman}, a high-quality dataset with realistic 
human-background interactions. Experiments demonstrate that InsHuman significantly reduces person count errors and achieves consistent improvements in pose adaptation, structural integrity, and facial identity preservation.

\newpage
\subsection*{AI use statement}
\lijie{In this work, we used generative AI tools for language polishing and editing of the manuscript. We used generative model for dataset construction and the evaluation of VLM-T metric.
We have reviewed all AI-assisted work and take responsibility for the final content of this work.}





\newpage
\bibliography{iclr2027_conference}

\begin{thebibliography}{51}
\providecommand{\natexlab}[1]{#1}
\providecommand{\url}[1]{\texttt{#1}}
\expandafter\ifx\csname urlstyle\endcsname\relax
  \providecommand{\doi}[1]{doi: #1}\else
  \providecommand{\doi}{doi: \begingroup \urlstyle{rm}\Url}\fi

\bibitem[Avrahami et~al.(2022)Avrahami, Lischinski, and Fried]{avrahami2022blended}
Omri Avrahami, Dani Lischinski, and Ohad Fried.
\newblock Blended diffusion for text-driven editing of natural images.
\newblock In \emph{Proceedings of the IEEE/CVF Conference on Computer Vision and Pattern Recognition}, pp.\  18208--18218, 2022.

\bibitem[Borse et~al.(2026)Borse, Choi, Park, Kim, Kadambi, Garrepalli, Yun, Malladi, and Porikli]{borse2026multihuman}
Shubhankar Borse, Seokeon Choi, Sunghyun Park, Jeongho Kim, Shreya Kadambi, Risheek Garrepalli, Sungrack Yun, Durga Malladi, and Fatih Porikli.
\newblock Multihuman-testbench: Benchmarking image generation for multiple humans.
\newblock \emph{Advances in Neural Information Processing Systems}, 38, 2026.

\bibitem[Brooks et~al.(2023)Brooks, Holynski, and Efros]{brooks2023instructpix2pix}
Tim Brooks, Aleksander Holynski, and Alexei~A Efros.
\newblock Instructpix2pix: Learning to follow image editing instructions.
\newblock In \emph{Proceedings of the IEEE/CVF Conference on Computer Vision and Pattern Recognition}, pp.\  18392--18402, 2023.

\bibitem[Cao et~al.(2018)Cao, Shen, Xie, Parkhi, and Zisserman]{cao2018vggface2}
Qiong Cao, Li~Shen, Weidi Xie, Omkar~M Parkhi, and Andrew Zisserman.
\newblock Vggface2: A dataset for recognising faces across pose and age.
\newblock In \emph{2018 13th IEEE international conference on automatic face \& gesture recognition (FG 2018)}, pp.\  67--74. IEEE, 2018.

\bibitem[Cao et~al.(2025)Cao, Chen, Chen, Cheng, Cui, Deng, Dong, Gong, Gu, Gu, et~al.]{cao2025hunyuanimage}
Siyu Cao, Hangting Chen, Peng Chen, Yiji Cheng, Yutao Cui, Xinchi Deng, Ying Dong, Kipper Gong, Tianpeng Gu, Xiusen Gu, et~al.
\newblock Hunyuanimage 3.0 technical report.
\newblock \emph{arXiv preprint arXiv:2509.23951}, 2025.

\bibitem[Chen et~al.(2024)Chen, Huang, Liu, Shen, Zhao, and Zhao]{chen2024anydoor}
Xi~Chen, Lianghua Huang, Yu~Liu, Yujun Shen, Deli Zhao, and Hengshuang Zhao.
\newblock Anydoor: Zero-shot object-level image customization.
\newblock In \emph{Proceedings of the IEEE/CVF Conference on Computer Vision and Pattern Recognition}, pp.\  6418--6427, 2024.

\bibitem[Choi et~al.(2021)Choi, Park, Lee, and Choo]{choi2021viton}
Seunghwan Choi, Sunghyun Park, Minsoo Lee, and Jaegul Choo.
\newblock Viton-hd: High-resolution virtual try-on via misalignment-aware normalization.
\newblock In \emph{Proceedings of the IEEE/CVF Conference on Computer Vision and Pattern Recognition}, pp.\  14131--14140, 2021.

\bibitem[Couairon et~al.(2023)Couairon, Verbeek, Schwenk, and Cord]{couairon2022diffedit}
Guillaume Couairon, Jakob Verbeek, Holger Schwenk, and Matthieu Cord.
\newblock Diffedit: Diffusion-based semantic image editing with mask guidance.
\newblock In \emph{International Conference on Learning Representations}, 2023.

\bibitem[Deng et~al.(2019)Deng, Guo, Xue, and Zafeiriou]{deng2019arcface}
Jiankang Deng, Jia Guo, Niannan Xue, and Stefanos Zafeiriou.
\newblock Arcface: Additive angular margin loss for deep face recognition.
\newblock In \emph{Proceedings of the IEEE/CVF conference on computer vision and pattern recognition}, pp.\  4690--4699, 2019.

\bibitem[Dinh et~al.(2022)Dinh, Nguyen, and Hua]{dinh2022tise}
Tan~M Dinh, Rang Nguyen, and Binh-Son Hua.
\newblock Tise: Bag of metrics for text-to-image synthesis evaluation.
\newblock In \emph{European Conference on Computer Vision}, pp.\  594--609. Springer, 2022.

\bibitem[Fu et~al.(2022)Fu, Li, Jiang, Lin, Qian, Loy, Wu, and Liu]{fu2022stylegan}
Jianglin Fu, Shikai Li, Yuming Jiang, Kwan-Yee Lin, Chen Qian, Chen-Change Loy, Wayne Wu, and Ziwei Liu.
\newblock Stylegan-human: A data-centric odyssey of human generation.
\newblock In \emph{European Conference on Computer Vision}, pp.\  1--19. Springer, 2022.

\bibitem[Gal et~al.(2023)Gal, Alaluf, Atzmon, Patashnik, Bermano, Chechik, and Cohen-Or]{gal2022image}
Rinon Gal, Yuval Alaluf, Yuval Atzmon, Or~Patashnik, Amit~H Bermano, Gal Chechik, and Daniel Cohen-Or.
\newblock An image is worth one word: Personalizing text-to-image generation using textual inversion.
\newblock In \emph{International Conference on Learning Representations}, 2023.

\bibitem[Gao et~al.(2025)Gao, Joseph, and De~La~Torre]{gao2025teleportraits}
Jialu Gao, KJ~Joseph, and Fernando De~La~Torre.
\newblock Teleportraits: Training-free people insertion into any scene.
\newblock In \emph{Proceedings of the IEEE/CVF International Conference on Computer Vision}, pp.\  18866--18875, 2025.

\bibitem[Hertz et~al.(2023)Hertz, Mokady, Tenenbaum, Aberman, Pritch, and Cohen-Or]{hertz2022prompt}
Amir Hertz, Ron Mokady, Jay Tenenbaum, Kfir Aberman, Yael Pritch, and Daniel Cohen-Or.
\newblock Prompt-to-prompt image editing with cross attention control.
\newblock In \emph{International Conference on Learning Representations}, 2023.

\bibitem[Huang et~al.(2023)Huang, Sun, Xie, Li, and Liu]{DBLP:conf/nips/HuangSXLL23}
Kaiyi Huang, Kaiyue Sun, Enze Xie, Zhenguo Li, and Xihui Liu.
\newblock T2i-compbench: {A} comprehensive benchmark for open-world compositional text-to-image generation.
\newblock In Alice Oh, Tristan Naumann, Amir Globerson, Kate Saenko, Moritz Hardt, and Sergey Levine (eds.), \emph{Advances in Neural Information Processing Systems 36: Annual Conference on Neural Information Processing Systems 2023, NeurIPS 2023, New Orleans, LA, USA, December 10 - 16, 2023}, 2023.
\newblock URL \url{http://papers.nips.cc/paper\_files/paper/2023/hash/f8ad010cdd9143dbb0e9308c093aff24-Abstract-Datasets\_and\_Benchmarks.html}.

\bibitem[Jiang et~al.(2022)Jiang, Yang, Qiu, Wu, Loy, and Liu]{jiang2022text2human}
Yuming Jiang, Shuai Yang, Haonan Qiu, Wayne Wu, Chen~Change Loy, and Ziwei Liu.
\newblock Text2human: Text-driven controllable human image generation.
\newblock \emph{ACM Transactions on Graphics (TOG)}, 41\penalty0 (4):\penalty0 1--11, 2022.

\bibitem[Ju et~al.(2023)Ju, Zeng, Zhao, Wang, Zhang, and Xu]{ju2023humansd}
Xuan Ju, Ailing Zeng, Chenchen Zhao, Jianan Wang, Lei Zhang, and Qiang Xu.
\newblock Humansd: A native skeleton-guided diffusion model for human image generation.
\newblock In \emph{2023 IEEE/CVF International Conference on Computer Vision (ICCV)}, pp.\  15942--15952. IEEE, 2023.

\bibitem[Karras et~al.(2019)Karras, Laine, and Aila]{karras2019style}
Tero Karras, Samuli Laine, and Timo Aila.
\newblock A style-based generator architecture for generative adversarial networks.
\newblock In \emph{Proceedings of the IEEE/CVF conference on computer vision and pattern recognition}, pp.\  4401--4410, 2019.

\bibitem[Kulal et~al.(2023)Kulal, Brooks, Aiken, Wu, Yang, Lu, Efros, and Singh]{kulal2023putting}
Sumith Kulal, Tim Brooks, Alex Aiken, Jiajun Wu, Jimei Yang, Jingwan Lu, Alexei~A Efros, and Krishna~Kumar Singh.
\newblock Putting people in their place: Affordance-aware human insertion into scenes.
\newblock In \emph{Proceedings of the IEEE/CVF Conference on Computer Vision and Pattern Recognition}, pp.\  17089--17099, 2023.

\bibitem[Kumari et~al.(2023)Kumari, Zhang, Zhang, Shechtman, and Zhu]{kumari2023multi}
Nupur Kumari, Bingliang Zhang, Richard Zhang, Eli Shechtman, and Jun-Yan Zhu.
\newblock Multi-concept customization of text-to-image diffusion.
\newblock In \emph{Proceedings of the IEEE/CVF Conference on Computer Vision and Pattern Recognition}, pp.\  1931--1941, 2023.

\bibitem[Labs(2025)]{flux2}
Black~Forest Labs.
\newblock {FLUX}.2.
\newblock Online. \url{https://blackforestlabs.ai}, 2025.
\newblock Accessed: 2025-05-07.

\bibitem[Li et~al.(2023)Li, Liu, Wu, Mu, Yang, Gao, Li, and Lee]{li2023gligen}
Yuheng Li, Haotian Liu, Qingyang Wu, Fangzhou Mu, Jianwei Yang, Jianfeng Gao, Chunyuan Li, and Yong~Jae Lee.
\newblock Gligen: Open-set grounded text-to-image generation.
\newblock In \emph{Proceedings of the IEEE/CVF Conference on Computer Vision and Pattern Recognition}, pp.\  22511--22521, 2023.

\bibitem[Li et~al.(2024)Li, Cao, Wang, Qi, Cheng, and Shan]{li2024photomaker}
Zhen Li, Mingdeng Cao, Xintao Wang, Zhongang Qi, Ming-Ming Cheng, and Ying Shan.
\newblock Photomaker: Customizing realistic human photos via stacked id embedding.
\newblock In \emph{Proceedings of the IEEE/CVF Conference on Computer Vision and Pattern Recognition}, pp.\  8640--8650, 2024.

\bibitem[Lin et~al.(2014)Lin, Maire, Belongie, Hays, Perona, Ramanan, Doll{\'a}r, and Zitnick]{lin2014microsoft}
Tsung-Yi Lin, Michael Maire, Serge Belongie, James Hays, Pietro Perona, Deva Ramanan, Piotr Doll{\'a}r, and C~Lawrence Zitnick.
\newblock Microsoft coco: Common objects in context.
\newblock In \emph{European conference on computer vision}, pp.\  740--755. Springer, 2014.

\bibitem[Liu et~al.(2016)Liu, Luo, Qiu, Wang, and Tang]{liu2016deepfashion}
Ziwei Liu, Ping Luo, Shi Qiu, Xiaogang Wang, and Xiaoou Tang.
\newblock Deepfashion: Powering robust clothes recognition and retrieval with rich annotations.
\newblock In \emph{Proceedings of the IEEE conference on computer vision and pattern recognition}, pp.\  1096--1104, 2016.

\bibitem[Lu et~al.(2023)Lu, Liu, and Kong]{lu2023tf}
Shilin Lu, Yanzhu Liu, and Adams Wai-Kin Kong.
\newblock Tf-icon: Diffusion-based training-free cross-domain image composition.
\newblock In \emph{2023 IEEE/CVF International Conference on Computer Vision (ICCV)}, pp.\  2294--2305. IEEE, 2023.

\bibitem[Mittal et~al.(2012)Mittal, Moorthy, and Bovik]{mittal2012no}
Anish Mittal, Anush~Krishna Moorthy, and Alan~Conrad Bovik.
\newblock No-reference image quality assessment in the spatial domain.
\newblock \emph{IEEE Transactions on image processing}, 21\penalty0 (12):\penalty0 4695--4708, 2012.

\bibitem[Munkres(1957)]{munkres1957algorithms}
James Munkres.
\newblock Algorithms for the assignment and transportation problems.
\newblock \emph{Journal of the society for industrial and applied mathematics}, 5\penalty0 (1):\penalty0 32--38, 1957.

\bibitem[Parihar et~al.(2024)Parihar, Gupta, VS, and Babu]{parihar2024text2place}
Rishubh Parihar, Harsh Gupta, Sachidanand VS, and R~Venkatesh Babu.
\newblock Text2place: Affordance-aware text guided human placement.
\newblock In \emph{European Conference on Computer Vision}, pp.\  57--77. Springer, 2024.

\bibitem[Peebles \& Xie(2023)Peebles and Xie]{peebles2023scalable}
William Peebles and Saining Xie.
\newblock Scalable diffusion models with transformers.
\newblock In \emph{Proceedings of the IEEE/CVF International Conference on Computer Vision}, pp.\  4199--4209, 2023.

\bibitem[Rombach et~al.(2022)Rombach, Blattmann, Lorenz, Esser, and Ommer]{rombach2022high}
Robin Rombach, Andreas Blattmann, Dominik Lorenz, Patrick Esser, and Bj{\"o}rn Ommer.
\newblock High-resolution image synthesis with latent diffusion models.
\newblock In \emph{Proceedings of the IEEE/CVF conference on computer vision and pattern recognition}, pp.\  10684--10695, 2022.

\bibitem[Ruiz et~al.(2023)Ruiz, Li, Jampani, Pritch, Rubinstein, and Aberman]{ruiz2023dreambooth}
Nataniel Ruiz, Yuanzhen Li, Varun Jampani, Yael Pritch, Michael Rubinstein, and Kfir Aberman.
\newblock Dreambooth: Fine tuning text-to-image diffusion models for subject-driven generation.
\newblock In \emph{Proceedings of the IEEE/CVF Conference on Computer Vision and Pattern Recognition}, pp.\  22500--22510, 2023.

\bibitem[Shao et~al.(2018)Shao, Zhao, Li, Xiao, Yu, Zhang, and Sun]{shao2018crowdhuman}
Shuai Shao, Zijian Zhao, Boxun Li, Tete Xiao, Gang Yu, Xiangyu Zhang, and Jian Sun.
\newblock Crowdhuman: A benchmark for detecting human in a crowd.
\newblock \emph{arXiv preprint arXiv:1805.00123}, 2018.

\bibitem[Song et~al.(2026)Song, Jiang, Yang, Cheng, Quan, and Yang]{song2026insert}
Wensong Song, Hong Jiang, Zongxin Yang, Zheqiao Cheng, Ruijie Quan, and Yi~Yang.
\newblock Insert anything: Image insertion via in-context editing in dit.
\newblock In \emph{Proceedings of the AAAI Conference on Artificial Intelligence}, volume~40, pp.\  9097--9105, 2026.

\bibitem[Song et~al.(2023)Song, Zhang, Lin, Cohen, Price, Zhang, Kim, and Ali]{song2023objectstitch}
Yizhi Song, Zhifei Zhang, Zhe Lin, Scott Cohen, Brian Price, Jianming Zhang, Seung~Yong Kim, and Daniel Ali.
\newblock Objectstitch: Object compositing with diffusion models.
\newblock In \emph{Proceedings of the IEEE/CVF Conference on Computer Vision and Pattern Recognition}, pp.\  18310--18319, 2023.

\bibitem[Temel et~al.(2016)Temel, Prabhushankar, and AlRegib]{DBLP:journals/spl/TemelPA16}
Dogancan Temel, Mohit Prabhushankar, and Ghassan AlRegib.
\newblock {UNIQUE:} unsupervised image quality estimation.
\newblock \emph{{IEEE} Signal Process. Lett.}, 23\penalty0 (10):\penalty0 1414--1418, 2016.
\newblock \doi{10.1109/LSP.2016.2601119}.
\newblock URL \url{https://doi.org/10.1109/LSP.2016.2601119}.

\bibitem[Tian et~al.(2025)Tian, Ye, and DOERMANN]{NEURIPS2025_71034442}
Yunjie Tian, Qixiang Ye, and DAVID DOERMANN.
\newblock Yolov12: Attention-centric real-time object detectors.
\newblock In D.~Belgrave, C.~Zhang, H.~Lin, R.~Pascanu, P.~Koniusz, M.~Ghassemi, and N.~Chen (eds.), \emph{Advances in Neural Information Processing Systems}, volume 38, Main Conference, pp.\  78433--78457. Curran Associates, Inc., 2025.
\newblock \doi{10.52202/085713-2627}.
\newblock URL \url{https://proceedings.neurips.cc/paper_files/paper/2025/file/7103444259031cc58051f8c9a4868533-Paper-Conference.pdf}.

\bibitem[Tu et~al.(2025)Tu, Luo, Chen, Ji, Bai, and Zhao]{tu2025videoanydoor}
Yuanpeng Tu, Hao Luo, Xi~Chen, Sihui Ji, Xiang Bai, and Hengshuang Zhao.
\newblock Videoanydoor: High-fidelity video object insertion with precise motion control.
\newblock In \emph{Proceedings of the Special Interest Group on Computer Graphics and Interactive Techniques Conference Conference Papers}, pp.\  1--11, 2025.

\bibitem[Tumanyan et~al.(2023)Tumanyan, Geyer, Bagon, and Dekel]{tumanyan2023plug}
Narek Tumanyan, Michal Geyer, Shai Bagon, and Tali Dekel.
\newblock Plug-and-play diffusion features for text-driven image-to-image translation.
\newblock In \emph{Proceedings of the IEEE/CVF Conference on Computer Vision and Pattern Recognition}, pp.\  1921--1930, 2023.

\bibitem[Wang et~al.(2023{\natexlab{a}})Wang, Chan, and Loy]{wang2023exploring}
Jianyi Wang, Kelvin~CK Chan, and Chen~Change Loy.
\newblock Exploring clip for assessing the look and feel of images.
\newblock In \emph{Proceedings of the AAAI conference on artificial intelligence}, volume~37, pp.\  2555--2563, 2023{\natexlab{a}}.

\bibitem[Wang et~al.(2024)Wang, Bai, Wang, Qin, Chen, Li, Tang, and Hu]{wang2024instantid}
Qixun Wang, Xu~Bai, Haofan Wang, Zekui Qin, Anthony Chen, Huaxia Li, Xu~Tang, and Yao Hu.
\newblock Instantid: Zero-shot identity-preserving generation in seconds.
\newblock \emph{arXiv preprint arXiv:2401.07519}, 2024.

\bibitem[Wang et~al.(2023{\natexlab{b}})Wang, Saharia, Montgomery, Pont-Tuset, Noy, Pellegrini, Onoe, Laszlo, Fleet, Soricut, et~al.]{wang2023imagen}
Su~Wang, Chitwan Saharia, Ceslee Montgomery, Jordi Pont-Tuset, Shai Noy, Stefano Pellegrini, Yasumasa Onoe, Sarah Laszlo, David~J Fleet, Radu Soricut, et~al.
\newblock Imagen editor and editbench: Advancing and evaluating text-guided image inpainting.
\newblock In \emph{2023 ieee/cvf conference on computer vision and pattern recognition (cvpr)}, pp.\  18359--18369. IEEE, 2023{\natexlab{b}}.

\bibitem[Wu et~al.(2025{\natexlab{a}})Wu, Li, Zhou, Lin, Gao, Yan, Yin, Bai, Xu, Chen, et~al.]{wu2025qwen}
Chenfei Wu, Jiahao Li, Jingren Zhou, Junyang Lin, Kaiyuan Gao, Kun Yan, Sheng-ming Yin, Shuai Bai, Xiao Xu, Yilei Chen, et~al.
\newblock Qwen-image technical report.
\newblock \emph{arXiv preprint arXiv:2508.02324}, 2025{\natexlab{a}}.

\bibitem[Wu et~al.(2025{\natexlab{b}})Wu, Zheng, Yan, Xiao, Luo, Wang, Li, Jiang, Liu, Zhou, et~al.]{wu2025omnigen2}
Chenyuan Wu, Pengfei Zheng, Ruiran Yan, Shitao Xiao, Xin Luo, Yueze Wang, Wanli Li, Xiyan Jiang, Yexin Liu, Junjie Zhou, et~al.
\newblock Omnigen2: Exploration to advanced multimodal generation.
\newblock \emph{arXiv preprint arXiv:2506.18871}, 2025{\natexlab{b}}.

\bibitem[Xia et~al.(2025)Xia, Peng, Zhang, Huang, Liu, Li, Tan, Wu, Wang, Wang, et~al.]{xia2025dreamomni2}
Bin Xia, Bohao Peng, Yuechen Zhang, Junjia Huang, Jiyang Liu, Jingyao Li, Haoru Tan, Sitong Wu, Chengyao Wang, Yitong Wang, et~al.
\newblock Dreamomni2: Multimodal instruction-based editing and generation.
\newblock \emph{arXiv preprint arXiv:2510.06679}, 2025.

\bibitem[Xiao et~al.(2023)Xiao, Yin, Freeman, Durand, and Han]{xiao2023fastcomposer}
Guangxuan Xiao, Tianwei Yin, William~T Freeman, Fr{\'e}do Durand, and Song Han.
\newblock Fastcomposer: Tuning-free multi-subject image generation with localized attention.
\newblock \emph{arXiv preprint arXiv:2305.10431}, 2023.

\bibitem[Xu et~al.(2024)Xu, Zhang, Liew, Yan, Liu, Zhang, Feng, and Shou]{DBLP:conf/cvpr/XuZLYLZFS24}
Zhongcong Xu, Jianfeng Zhang, Jun~Hao Liew, Hanshu Yan, Jia{-}Wei Liu, Chenxu Zhang, Jiashi Feng, and Mike~Zheng Shou.
\newblock Magicanimate: Temporally consistent human image animation using diffusion model.
\newblock In \emph{{IEEE/CVF} Conference on Computer Vision and Pattern Recognition, {CVPR} 2024, Seattle, WA, USA, June 16-22, 2024}, pp.\  1481--1490. {IEEE}, 2024.
\newblock \doi{10.1109/CVPR52733.2024.00147}.
\newblock URL \url{https://doi.org/10.1109/CVPR52733.2024.00147}.

\bibitem[Yang et~al.(2023)Yang, Gu, Zhang, Zhang, Chen, Sun, Chen, and Wen]{yang2023paint}
Binxin Yang, Shuyang Gu, Bo~Zhang, Ting Zhang, Xuejin Chen, Xiaoyan Sun, Dong Chen, and Fang Wen.
\newblock Paint by example: Exemplar-based image editing with diffusion models.
\newblock In \emph{Proceedings of the IEEE/CVF Conference on Computer Vision and Pattern Recognition}, pp.\  18381--18391, 2023.

\bibitem[Ye et~al.(2023)Ye, Zhang, Liu, Han, and Yang]{ye2023ip}
Hu~Ye, Jun Zhang, Sibo Liu, Xiao Han, and Wei Yang.
\newblock Ip-adapter: Text compatible image prompt adapter for text-to-image diffusion models.
\newblock \emph{arXiv preprint arXiv:2308.06721}, 2023.

\bibitem[Zhang et~al.(2026)Zhang, Wang, Tang, and Wang]{zhang2026insert}
Yifan Zhang, Jianguo Wang, Zhongliang Tang, and Wenmin Wang.
\newblock Insert anyone: High-fidelity full-body photo insertion via dual-branch adapters.
\newblock \emph{Expert Systems with Applications}, pp.\  131013, 2026.

\bibitem[Zhu et~al.(2022)Zhu, Wayne, Qiu, Zhu, et~al.]{zhu2022celebv}
Hao Zhu, Wu~Wayne, Wentao Qiu, Chenxia Zhu, et~al.
\newblock Celebv-hq: A large-scale video facial attributes dataset.
\newblock In \emph{European conference on computer vision}, pp.\  650--667. Springer, 2022.

\end{thebibliography}
\bibliographystyle{iclr2027_conference}

\newpage
\appendix
\section*{Appendix}
\label{sec:Appendix}
\section{Implementation details}
\label{supp:imp_detail}
\noindent\textbf{Training Configuration.} We fine-tune Qwen-Image-Edit-2509~\citep{wu2025qwen} using LoRA with the proposed \emph{HBAF} and \emph{FFIP} strategies. Training is conducted on 2 NVIDIA A100 GPUs for a total of 2,645 iterations. The batch size is set to 2 and the learning rate is $1 \times 10^{-4}$. We use the AdamW optimizer with a weight decay of 0.01. For HBAF, we set $T_{\text{start}}$ to 900 and $T_{\text{end}}$ to 808. For FFIP, the face loss balancing coefficient $\lambda_{face}$ is set to 0.02.

\noindent\textbf{Inference Configuration.} All inference experiments are conducted on a single NVIDIA A100 GPU. Generating one image at a resolution of 1312×864 requires 40 denoising steps, with a total runtime of approximately 2.5 minutes, corresponding to an average of 3.75 seconds per step.


\section{Evaluation Details}
\label{supp:metrics}
\noindent\textbf{IDS (Identity Similarity).} 
Following MultiHuman-Testbench~\citep{borse2026multihuman}, we evaluate facial identity consistency using the IDS metric. We employ the InsightFace library to detect all faces in the generated image and the reference image, extracting normalized feature vectors for each. We then compute the cosine similarity matrix between all detected faces across the two images and solve for the optimal one-to-one face matching via the Hungarian algorithm~\citep{munkres1957algorithms}. The IDS score is defined as: $\textit{IDS} = S_{total} / \max(n_{gen}, n_{src})$. $S_{total}$ is the sum of cosine similarities over all matched face pairs, and $n_{gen}$ and $n_{src}$ denote the number of detected faces in the generated and reference images, respectively. 

\noindent\textbf{LPIPS.} Following Text2Place~\citep{parihar2024text2place}, we use LPIPS to measure how well the generated image preserves the target background. LPIPS computes a perceptual distance between the background regions of the synthesized and target images in feature space; a lower value indicates better background fidelity.

\noindent\textbf{CA (Counting Alignment).} Following the CA (Counting Alignment) metric proposed in TISE~\citep{dinh2022tise}, we adopt the number of persons in the reference image as the ground truth and compute the mean absolute error between the detected person count in the synthesized image and the reference, measuring whether the correct number of people is preserved after portrait insertion. A lower CA indicates better counting accuracy.

\noindent\textbf{CLIP-IQA, BRISQUE, and UNIQUE.} We employ CLIP-IQA~\citep{wang2023exploring}, BRISQUE~\citep{mittal2012no}, and UNIQUE~\citep{DBLP:journals/spl/TemelPA16} to evaluate the perceptual quality of the synthesized images, which can indirectly reflect the quality of the portrait insertion task's completion.

\noindent\textbf{VLM-T.} Following TelePortraits~\citep{gao2025teleportraits}, we use the VLM-T metric to evaluate the overall quality of human insertion. The full prompt used for evaluation is provided in List~\ref{lst:vlm-t_prompt} (Appendix~\ref{supp:prompts}).


\noindent\textbf{Compared Methods.} We compare against accessible state-of-the-art image editing models and 
dedicated insertion model, including DreamOmni2~\citep{xia2025dreamomni2}, HunyuanImage-3.0-Instruct~\citep{cao2025hunyuanimage}, OmniGen2~\citep{wu2025omnigen2}, Insert-Anything~\citep{song2026insert}, Nano-Banana-Pro and Qwen-Image-Edit-2509~\citep{wu2025qwen}.

\section{Additional Ablation Results}
\label{supp:ablation}
We present supplementary ablation experiments result to further justify the design choices of InsHuman, including the effect of the $\lambda_{\max}$ and $\lambda_{t}$ in HBAF and the mask design strategy.
\subsection{Effect of $\lambda_{\max}$ in HBAF}
\label{supp:ablation-lambda_max}
\begin{figure}[t]
  \centering
  \includegraphics[width=\textwidth]{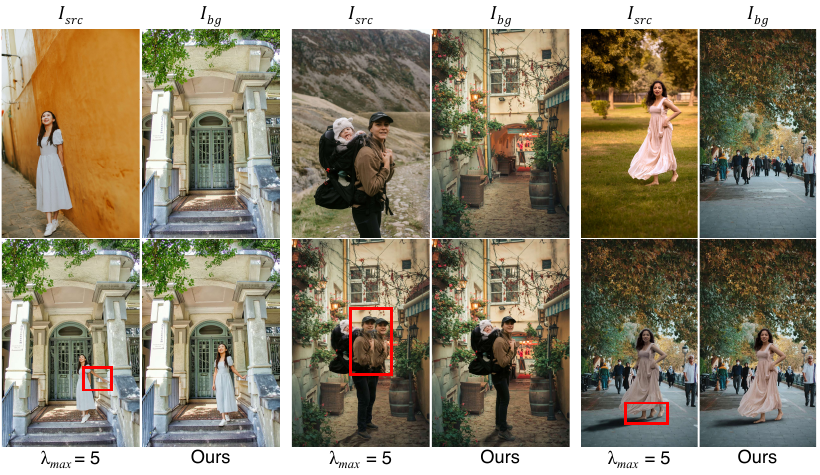}
  \caption{{\bf Visual comparison with \bm{$\lambda_{\max}$} = 5.} An excessively large value of $\lambda_{\max}$ will disrupt the generative process and produce anatomically implausible outputs.}
  \label{fig:ablation-λ_max}
\end{figure}

Setting $\lambda_{\max}=5$ imposes overly strong supervision on the human region during the early denoising stages, where the model is still forming global composition. As shown in Figure~\ref{fig:ablation-λ_max}, this results in severe structural artifacts, including phantom limbs, duplicated heads, and distorted body proportions. The CA metric correspondingly degrades to 0.2824, significantly worse than our default ($\lambda_{\max}=2.5$, CA$=$0.1298). This demonstrates that while amplifying the human-region weight can in principle strengthen structural constraints, an excessively large value disrupts the generative process and produces anatomically implausible outputs. A moderate $\lambda_{\max}=2.5$ strikes the right balance between structural guidance and natural generation.

\subsection{Effect of the Dynamic Weight Schedule $\lambda_t$ in HBAF}
\label{supp:ablation-fix_lambda}
\begin{figure}[t!]
  \centering
  \includegraphics[width=\textwidth]{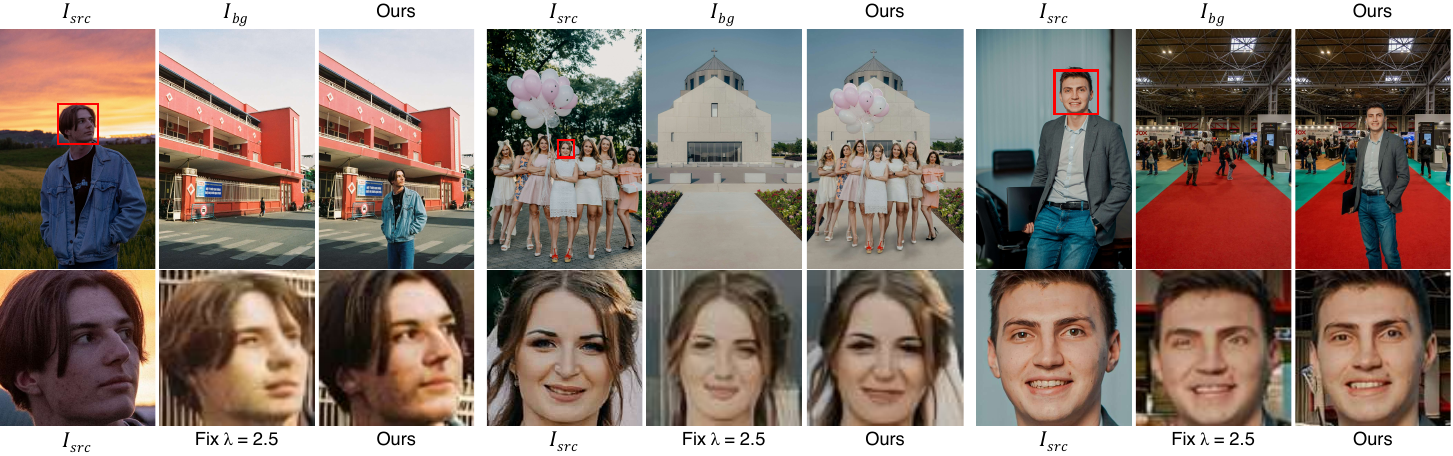}
  \caption{{\bf Visual comparison with a fixed \bm{$\lambda$}.} Removal of the dynamic weight function $\lambda_{t}$ disrupts facial feature generation, leading to identity deviations from the reference image.}
  \label{fig:ablation-fixλ}
\end{figure}

When the dynamic schedule $\lambda_t$ is replaced by a constant $\lambda=2.5$ applied uniformly across all timesteps, as shown in Figure~\ref{fig:ablation-fixλ}, this causes noticeable facial feature deviations from the source identity. 
Specifically, applying a constant strong constraint throughout all denoising stages limits the model’s ability to refine identity-related details in the later denoising steps, resulting in noticeable facial feature deviations from the reference image. In contrast, our dynamic weighting strategy assigns stronger guidance at early timesteps, when the model mainly determines global human structure and spatial arrangement, and gradually reduces the constraint during later timesteps to allow fine-grained facial features refinement. This adaptive schedule enables the model to simultaneously maintain reasonable human-background interaction and preserve facial identity consistency.

\subsection{Comparison of HBAF Mask Designs}
\label{supp:ablation-SAMBbox}
\begin{figure}[!htbp]
    \centering
    \includegraphics[width=\textwidth]{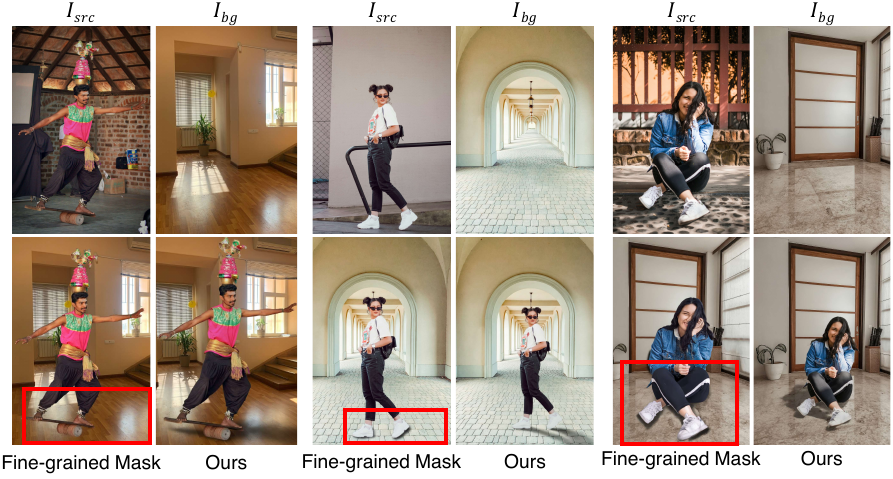}
    \caption{\lijie{\bf Comparison of different HBAF mask designs.} \lijie{The box-based mask (Ours) provides a broader supervision region around the human and yields more natural human-background transitions than the fine-grained mask, which excluding the boundary areas.}}
\label{fig:supp-sam-bbox}
\end{figure}
\lijie{We further investigate the effect of different mask designs on human insertion performance in HBAF. In our original implementation, HBAF first employs YOLOv12-m~\citep{NEURIPS2025_71034442} to detect foreground persons and merges their bounding boxes into a binary human-region mask. To construct a more fine-grained alternative, we retain the same YOLOv12-m bounding boxes and further apply SAM within each bounding box to segment the precise human foreground.}

\lijie{As shown in Figure~\ref{fig:supp-sam-bbox}, the box-based mask used in our method produces more natural transitions between the inserted person and the background. This is because natural visual blending in human insertion depends not only on the appearance of the person itself, but also on the transition and interaction regions surrounding the person, such as contact boundaries with the ground, cast shadows, and local illumination transitions. Therefore, a broader supervision region helps the model better learn these transition regions.}


\section{Image Sources}
\label{supp:Image-Sources}
For the real images used in constructing our dataset, we obtained the images directly from Unsplash (\url{https://unsplash.com/}) and used them in accordance with the applicable Unsplash License and Terms of Service.

\section{More qualitative comparisons with image editing models}
\label{supp:visual-comparison}
\vspace{-2ex}
\begin{figure}[!htbp]
    \centering
    \includegraphics[width=\textwidth]{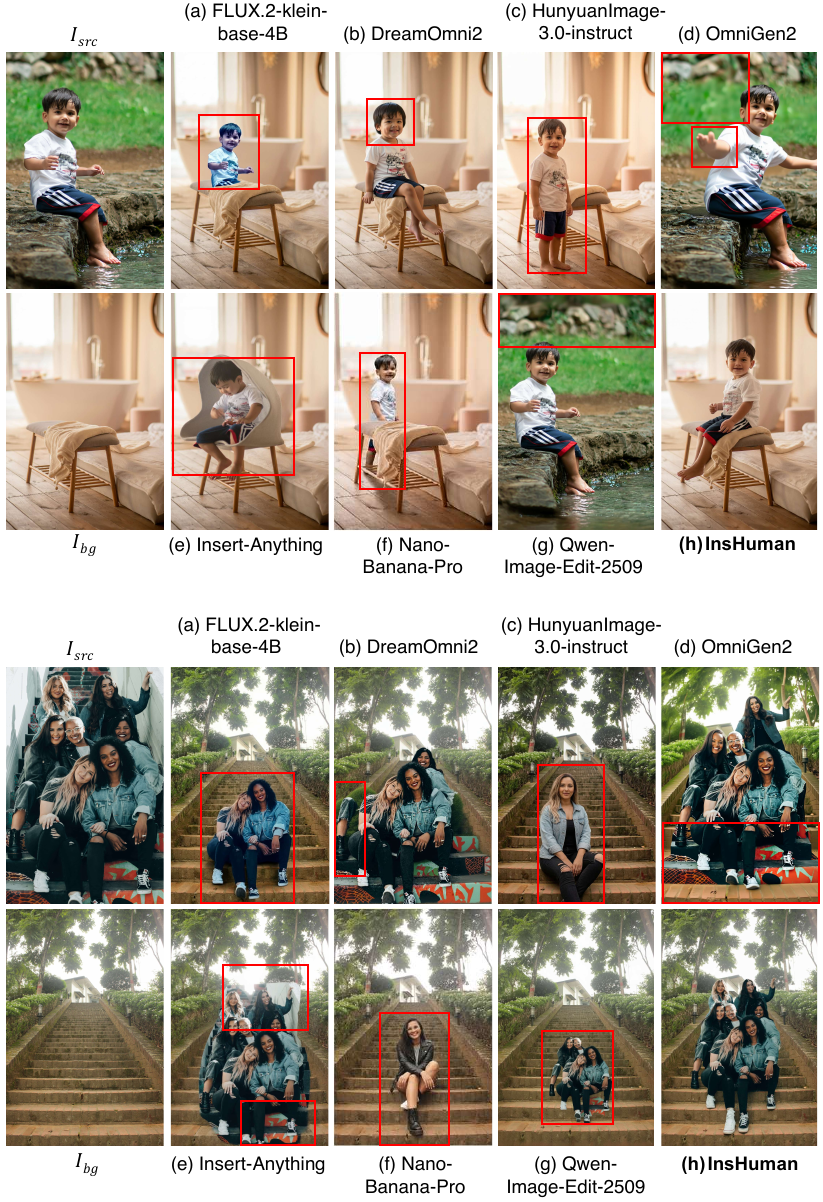}
\label{fig:supp1}
\end{figure}
\vspace{-2ex}
\begin{figure}[!htbp]
    \centering
    \includegraphics[width=0.93\textwidth]{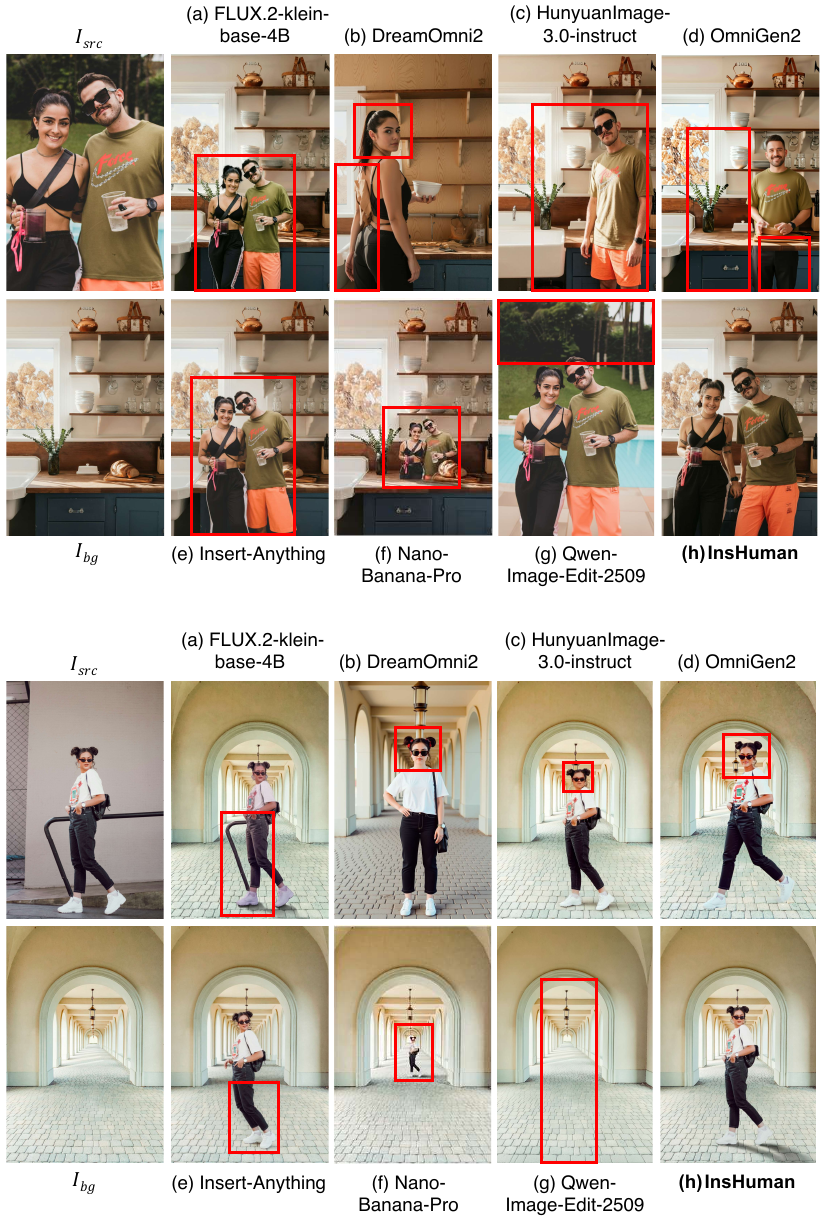}
    \caption{{\bf More qualitative comparison of different image editing models on human insertion.} Existing image editing models often exhibit issues on human insertion, e.g., poses failing to adapt to the background, deviations in the number of people or overall appearance from the reference image, and loss of facial features. In contrast, our method (InsHuman) more stably maintain human structure and identity features, achieving more natural human insertion. The \textbf{\textcolor{redbox}{red}} boxes highlight the errors.}
\label{fig:supp1}
\end{figure}

\newpage
\section{\lijie{Prompts used}}
\label{supp:prompts}
\begin{lstlisting}[
    caption={Nano Banana Pro prompts used for dataset synthesis.},
    label={lst:nano_prompt}
]
Task: Precisely place the figure in the foreground of [the first image] into the background scene of [the second image].
Requirements: 
1. Ensure the correct size and perspective. You may adjust the pose and position of the figure as appropriate to ensure a perfect match with the ground and objects in the background. 
2. Identity Constraint: Strictly maintain the facial features, clothing (including color and shape), and identity of the figure. You may generate additional information such as arms or legs, but must not alter the identity or micro-expressions of the figure. Also, ensure overall clarity, especially of the face. 
3. Environmental Integration: Achieve perfect geometric consistency and natural integration of lighting and shadows with the background scene.
\end{lstlisting}

\begin{lstlisting}[
    caption={Qwen-Image-Edit-2509 prompts used for inpainting.},
    label={lst:inapaint_prompt}
]
Remove the person from the scene. 
Fill the area naturally with the surrounding background, keeping lighting, shadows, textures, and perspective consistent. Seamless, photorealistic result with no traces of the removed subject.
\end{lstlisting}

\begin{lstlisting}[
    caption={Qwen2.5-VL-72B-Instruct prompts used for evaluating VLM-T.},
    label={lst:vlm-t_prompt}
]
### Task Definition
You will be provided with three images: the first is a composite image where one or more persons from a foreground image have been inserted into a background scene; the second is the foreground image containing the original person(s); the third is the original background scene image.

As an experienced evaluator, your task is to evaluate how well the insertion task has been completed, focusing on person count accuracy, scale, perspective, natural placement, and environmental integration.

### Scoring Criteria
Evaluate how well the composite image fulfills the following five requirements:

1. Person Count Accuracy: Check whether the number of persons visible in the composite image matches the number of persons in the foreground image. Missing or extra persons should significantly lower the score.

2. Scale and Perspective: Assess whether the size of the inserted person(s) is consistent with the background scene's perspective and depth, appearing proportionally correct and natural relative to surrounding objects.

3. Pose and Position Naturalness: Evaluate whether the pose and position of the inserted person(s) are natural and appropriate for the background context (e.g., feet properly grounded, body correctly oriented relative to objects in the scene).

4. Environmental Integration: Assess whether lighting direction, shadow quality, and geometric consistency between the inserted person(s) and the background are realistic and natural. Look for signs of unnatural edges, color mismatches, or lighting inconsistencies.

5. Overall Realism: Evaluate the overall visual coherence and photorealism of the composite image as a whole.

### Scoring Range
Give a specific integer score from 0 to 9:
- Very Poor (0): Task completely failed. No person detected in the composite image, or the number of persons is completely incorrect.
- Poor (1-2): Major failures. Significant issues in person count, scale, placement, or integration. Result looks clearly and unnaturally composited.
- Fair (3-4): Partial success. Some requirements met but with notable issues in at least two of the five criteria.
- Good (5-6): Most requirements met. Only minor issues in scale, perspective, or environmental integration.
- Very Good (7-8): Nearly all requirements met. Only minor imperfections remain, such as slight lighting inconsistency or minor scale deviation.
- Excellent (9): All requirements perfectly met. The composite is photorealistic, natural, and indistinguishable from a real photograph.

### Input Format
You will receive three images: the first is the composite (generated) image, the second is the foreground reference image, the third is the original background scene image.

### Output Format
[Your Score]
Output only the integer score. Do not include any analysis or explanation.
\end{lstlisting}





\end{document}